\documentclass[10pt,twocolumn,letterpaper]{article}
\usepackage{graphicx}
\usepackage{wrapfig}
\usepackage{subcaption}
\usepackage{multirow}
\usepackage{booktabs}
\usepackage{makecell}
\usepackage{pifont}
\usepackage{amsfonts}
\usepackage{xcolor}
\usepackage{xspace}
\usepackage[accsupp]{axessibility}
\usepackage[table]{xcolor}
\definecolor{lightgray}{gray}{0.92} 

\newcommand{\cmarkg}{\ding{51}\xspace}%
\usepackage[pagenumbers]{wacv} 

\definecolor{wacvblue}{rgb}{0.21,0.49,0.74}
\usepackage[pagebackref,breaklinks,colorlinks,allcolors=wacvblue]{hyperref}

\def\wacvPaperID{75} 
\def\confName{WACV}
\def\confYear{2027}

\title{TIE: Training-free CLIP Embedding Guidance for Compositional \\ Text-to-Image Alignment}

\author{
Sujung Hong\thanks{Equal contribution.}\hspace{1em}
Tae Eun Choi$^{*}$\hspace{1em}
Youngjun Jun\hspace{1em}
Chanyong Yoon\hspace{1em}
Seong Jae Hwang\thanks{Corresponding author.}\\[0.5em]
Yonsei University\\[0.3em]
{\tt\small \{sujung0914, teunchoi, youngjun, charlie2019370, seongjae\}@yonsei.ac.kr}
}

\begin{document}
\maketitle
\begin{abstract}
Text-to-image (T2I) diffusion models have advanced notably, but still fail to satisfy prompt conditions, resulting in attribute mismatches and missing objects.
These errors normally fall into two categories: \textit{concept loss}, when an object or attribute (i.e., concept) goes missing, and \textit{concept confusion}, when an attribute is assigned to the wrong object or multiple objects blend together.
Recent studies suggest that these challenges are caused by the limited capacity of CLIP text encoder to capture fine-grained semantic details. 
Although several methods have been proposed, concept loss and concept confusion persist in recent T2I models when handling multi-object, multi-attribute prompts. 
In this paper, we conduct an embedding-level analysis to develop a
training-free, inference-time solution for concept loss and concept confusion
across CLIP-based T2I models.
We find that (1) concepts mentioned later exhibit higher embedding entropy,
reflecting less concentrated representations that are more often associated
with concept loss, and (2) the first CLIP attention layer captures the
strength of binding between each object and its attribute.
Guided by our findings, we introduce TIE, a method that improves semantic alignment through a single text-embedding update. TIE addresses concept loss via entropy-aware singular value amplification and resolves concept confusion through interpolation–extrapolation binding based on CLIP attention scores, all in a training-free manner.
Extensive experiments demonstrate that TIE enhances semantic fidelity in multi-concept scenarios with minimal sampling overhead.
The project page is available at \url{https://hong-sujung.github.io/TIE/}.
\end{abstract}
    
\vspace{-10pt}
\section{Introduction}
\label{sec:introduction}

\begin{figure}
  \centering
  \includegraphics[width=\linewidth]{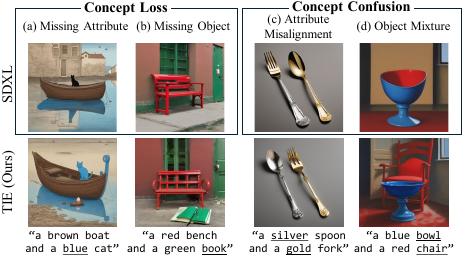}
  \caption{%
    \textbf{Failure cases in simple prompts.}
  }
  \label{fig:fig1_overview2}
  \vspace{-20pt}
\end{figure}

Text-to-image (T2I) generation has advanced substantially through the development of diffusion models~\cite{sdxl, sd,esser2024scaling, imagen}, achieving notable performance in generating images from general prompts~\cite{sd, imagen, ramesh2022hierarchical}.
However, they continue to struggle with prompts involving multiple objects and attributes (e.g., color, shape), often resulting in semantic inconsistencies~\cite{ramesh2022hierarchical,clark2023text}. 
For example, given a simple prompt \texttt{"a brown boat and a blue cat"}, the generated image sometimes contains only a brown boat, or falsely depicts a brown cat and a blue boat. 
As prompt complexity increases, such failures occur more frequently and manifest in more varied forms.

In text-to-image (T2I) synthesis, the text prompt itself specifies the desired constraints on the output, in which objects should appear and which attributes should be bound to each object.
Failures in controlled generation can therefore be viewed as violations of these prompt-induced constraints.
For clear understanding, these failures can be classified into two categories: \textit{concept loss} and \textit{concept confusion}, as illustrated in Fig.~\ref{fig:fig1_overview2}.
\textit{Concept loss} occurs when some of the intended objects or attributes specified in the prompt are missing from the generated image.  
This concept loss includes cases where objects appear without their corresponding attributes (Fig.~\ref{fig:fig1_overview2}a) or where an object is entirely omitted (Fig.~\ref{fig:fig1_overview2}b). 
In contrast, \textit{concept confusion} arises when the binding between attributes and objects are incorrectly aligned.
Such misalignment leads attributes to be assigned to unrelated objects (Fig.~\ref{fig:fig1_overview2}c) or merges objects (Fig.~\ref{fig:fig1_overview2}d).
These failures indicate that even large-scale T2I models still struggle with multi-object and multi-attribute prompts.

Several studies~\cite{sdxl, ane, sdg} have attributed these
failures to limitations in the textual representations used by T2I models.
Although the widely used CLIP text encoder~\cite{radford2021learning}
provides rich textual representations, it often struggles to capture
fine-grained details~\cite{bow,cab,zarei2025improvingcompositionalattributebinding,
jing2024fineclip}.
To address this limitation, previous works have proposed various strategies
for improving compositional generation.
Some methods directly manipulate text representations by merging
tokens~\cite{tome}, adaptively binding embeddings~\cite{magnet}, or
iteratively optimizing them~\cite{chen2024cat}, while others rely on
additional supervision such as prompt parsing~\cite{sdg}.
More recent approaches instead intervene during diffusion sampling, for
example by rebalancing cross-modal attention~\cite{lv2025rethinking} or steering
flow-matching dynamics~\cite{bill2026focus}.
However, these methods either target specific compositional failures, require
additional supervision, or introduce extra sampling-time computation.
In this work, we perform a concept-level analysis of the CLIP text encoder
and its embeddings to devise a unified and efficient solution to both concept
loss and concept confusion.

To address these challenges, we identify two indicators of concept loss and concept confusion.
First, to address concept loss, we observe a positional bias in which concepts appearing later in the prompt are more likely to be omitted in the generated image.
Based on this finding, we look for a quantitative measure which estimates the likelihood of concept loss.
Through various observations, we find that \textit{token entropy},
the entropy of each token embedding, reflects how diffusely its information
is distributed and serves as a proxy of concept loss.
Second, to mitigate concept confusion, we look into the first layer attention scores of the CLIP text encoder, referred to as \textit{L0 attention scores}. 
We demonstrate that L0 attention scores effectively show the strength of binding between concept tokens and how much it should be strengthened to reduce concept confusion. 
These two indicators are associated with \textit{both} concept loss and concept confusion using only the internal mechanisms of the CLIP text encoder.

Based on our findings, we propose two methods to address the limitations of T2I models: Adaptive Embedding Preservation (AEP) for mitigating concept loss, and Interpolation-Extrapolation Binding (IEB) for resolving concept confusion.
AEP adaptively amplifies tokens according to token entropy to encourage their presence in the generated image.
IEB enhances the semantic binding strength between objects and their corresponding attributes based on the L0 layer, while simultaneously suppressing the binding strength of irrelevant attribute-object pairs.
We refer to the combined framework of AEP and IEB as \textbf{TIE}. 
TIE improves text-image semantic alignment across multi-attribute and multi-object scenarios.
Our method updates text embeddings in a single step without training,
making it readily applicable to existing CLIP-based T2I models while
improving semantic fidelity.

Our main contributions are as follows:
\begin{itemize}
    \item We identify two indicators that jointly address concept loss and concept confusion.
    First, token entropy captures how concentrated each token representation is
    and provides a useful cue for mitigating concept loss.
    Secondly, the L0 attention score effectively captures attribute-object binding and serves as a indicator for how much the binding should be adjusted to address concept confusion.
    \item We propose TIE, inference-time framework that enhances text-image alignment via a single-step embedding update using AEP and IEB.
    AEP amplifies concept tokens with high embedding entropy to prevent concept loss, while IEB adjusts inter-concept relations using CLIP attention to clarify associations between each object and its attribute, thereby addressing concept confusion.
    \item Experimentally, TIE improves prompt-constraint satisfaction on multi-attribute and multi-object benchmarks with minimal inference-time overhead compared to baselines.
\end{itemize}

\section{Related Works}
\label{sec:relatedworks}
Text-to-image (T2I) diffusion models have advanced remarkably, enabling the generation of realistic images from a given prompt~\cite{sdxl,sd,esser2024scaling,controlnet}. 
To generate text conditional images, T2I diffusion models integrate embeddings from a text encoder into the denoising network through mechanisms such as cross-attention~\cite{imagen,ramesh2022hierarchical,gligen,glide,pixarta}.
However, they often struggle to preserve the semantic alignment between prompts and generated images when multiple objects or attributes are involved~\cite{ramesh2022hierarchical, clark2023text, dalle2fail, dalle2double}.
\vspace{-10pt}
\paragraph{Inference-time Methods.}
Prior methods improve compositional generation by intervening in the diffusion sampling process.
A\&E~\cite{ane} pioneered sampling-time cross-attention modification, while SynGen~\cite{syngen} and EBAMA~\cite{ebama} enhance concept binding through attention alignment and energy-based guidance.
InitNO~\cite{initno} instead optimizes the initial noise, and CONFORM~\cite{conform} employs a contrastive objective to improve generation fidelity.
More recently, TACA~\cite{lv2025rethinking} rebalances cross-modal attention in MM-DiTs, while FOCUS~\cite{bill2026focus} steers flow-matching dynamics through stochastic optimal control.
Despite their effectiveness, these methods incur additional sampling-time computation, limiting their practical efficiency.

\begin{figure*}[t!]
  \centering
  \includegraphics[width=\textwidth]{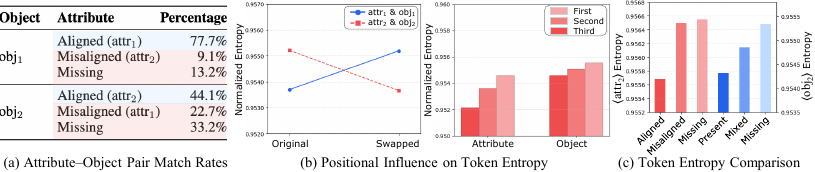} 
  \caption{
    \textbf{Analysis of positional bias and token entropy on concept loss.}
    (a) When issuing the prompt \texttt{"a ⟨attr$_1$⟩ ⟨obj$_1$⟩ and a ⟨attr$_2$⟩ ⟨obj$_2$⟩"}, \texttt{⟨attr$_2$⟩} and \texttt{⟨obj$_2$⟩} is far more prone to being misaligned or missing.
    (b) Left: The attribute-object pair mentioned later in the prompts shows higher entropy than the pair mentioned earlier, regardless of the tokens themselves. Right: Same holds for 3-pair prompts. 
    (c) Comparison of entropy for \texttt{⟨attr$_2$⟩} (red) and \texttt{⟨obj$_2$⟩} (blue) tokens each for when \texttt{⟨attr$_2$⟩} is Aligned, Misaligned or Missing, and when \texttt{⟨obj$_2$⟩} is Present, Mixed, or Missing, demonstrating that higher entropy is associated with concept loss.
   }
  \label{fig:fig2_analysis1}
\vspace{-10pt}
\end{figure*}
\vspace{-5pt}
\paragraph{Text Embedding-based Methods.}
Another line of work aims to improve the CLIP text embeddings~\cite{radford2021learning}, which is widely used but tends to capture broad semantic meanings while missing fine-grained details~\cite{bow, cab}. 
In addition, it is necessary to address the problem that the CLIP text encoder is biased toward objects mentioned earlier in a sentence~\cite{clipmulti,cliphighresolution}.
Magnet~\cite{magnet} explores modifying only the text embeddings to enhance semantic coherence. ToMe~\cite{tome} and TeeMo~\cite{ teemo} further refine embedding representations by investigating token merging and geometric properties, respectively. However, such embedding-based approaches tend to show weaker text-image alignment compared to cross-attention-based ones. 
In this paper, we analyze the characteristics of the commonly used CLIP text encoder and its embeddings, and explore the potential of embedding-based methods.
\vspace{-5pt}
\paragraph{RL-based Post-training Methods.}
A recent line of work improves semantic binding by post-training the generative model itself with reinforcement learning~\cite{Flow-GRPO, DanceGRPO, Expand-and-Prune}. 
These methods apply Group Relative Policy Optimization (GRPO)-style optimization to diffusion and flow-matching backbones, directly optimizing text--image alignment with reward signals and achieving strong compositional generation performance. 
They operate on the denoising network while keeping the text encoder fixed, whereas our method modifies the text conditioning itself without any training. 
Such reward-based post-training and our approach therefore address different stages of the generation pipeline and can be applied in a complementary manner in future work.

\section{Analysis of Concept Loss and Confusion}
\label{sec:analysis}

CLIP text encoders~\cite{radford2021learning}, widely used in
text-to-image (T2I) models, provide rich textual representations but often
fail to preserve fine-grained compositional information~\cite{bow,cab}.
Such failures can lead to \emph{concept loss}, where a requested concept is
not faithfully rendered, and \emph{concept confusion}, where attributes are
associated with the wrong objects~\cite{
zarei2025improvingcompositionalattributebinding,jing2024fineclip,conform}.
To understand these failures within the text encoder, we identify two internal
signals: \emph{token entropy} as a proxy for concept-loss propensity
(Section~\ref{sec:analysis1}) and \emph{L0 attention} as a signal of
attribute--object binding strength (Section~\ref{sec:analysis2}).
For controlled analysis, we first study simple multi-attribute prompts.
Because both quantities are computed solely from text-encoder representations,
the same measurements can be applied to longer and syntactically more complex
prompts.

\paragraph{Preliminary.}
T2I diffusion models~\cite{sdxl,sd} typically consist of a text encoder and a
denoising network.
We focus on models using the CLIP text encoders adopted in the Stable
Diffusion family~\cite{sdxl,sd,esser2024scaling}.
Given an input prompt $\mathcal{P}$, the CLIP text encoder $\tau$ produces
text embeddings under causal self-attention, such that token $i$ attends only
to tokens at positions $j \leq i$.
The resulting representation is
$C=\tau(\mathcal{P})\in\mathbb{R}^{N\times M}$, where $N$ is the number of
tokens and $M$ is the embedding dimension.
Padding is used to obtain a fixed sequence length.

\subsection{Token Entropy and Concept Loss}
\label{sec:analysis1}

Understanding which text representations are vulnerable to concept loss is a
necessary first step toward correcting them.
Previous work~\cite{chen2024cat} shows that objects mentioned later in a prompt
are more likely to disappear from generated images, revealing a positional bias
associated with CLIP's causal text encoding.
We first examine whether a similar positional effect occurs for attributes and
then ask whether an internal property of the corresponding token embeddings can
serve as a proxy for such failures.

\paragraph{Positional Bias in Attribute Tokens.}
We sample prompts of the form
\texttt{"a $\langle$attr$_1\rangle$ $\langle$obj$_1\rangle$ and a
$\langle$attr$_2\rangle$ $\langle$obj$_2\rangle$"}
from Concept Conjunction 500 (CC-500)~\cite{sdg} and retain 220 generated
images in which both objects are present.
For each image, we manually categorize the attribute associated with each
object as \emph{Aligned}, \emph{Misaligned}, or \emph{Missing}.
As shown in Fig.~\ref{fig:fig2_analysis1}a, the later-mentioned
attribute--object pair exhibits substantially more missing and misaligned
attributes than the earlier pair.
Thus, the positional bias previously observed for objects also manifests in
attribute rendering.
This observation alone, however, cannot determine whether a particular prompt
is prone to concept loss.
We therefore examine the geometry of individual token embeddings.

\vspace{-5pt}
\paragraph{Token Entropy and Embedding Geometry.}
For a token embedding $c_i\in\mathbb{R}^{d}$, we define normalized token
entropy as
{\small
\begin{equation}
\label{eq:token_entropy}
p_{i,j}
=
\frac{|c_{i,j}|}
{\sum_{k=1}^{d}|c_{i,k}|},
\qquad
\mathcal{H}_i
=
-\frac{1}{\log d}
\sum_{j=1}^{d}
p_{i,j}\log p_{i,j}.
\end{equation}}
This quantity is scale-invariant and increases as the representation is distributed across a larger effective number of coordinates, a notion closely related to effective dimensionality~\cite{roy2007effective}.
We further find that this dispersion has a concrete geometric interpretation
in CLIP.
CLIP representations are anisotropic and share a dense mean embedding direction $\hat{\mu}=\mu/\|\mu\|_2$~\cite{ethayarajh2019contextual}.
Token entropy strongly correlates with the cosine alignment between $c_i$ and $\hat{\mu}$ (Spearman $|\rho|=0.88$ for CLIP-G), suggesting that higher entropy reflects stronger alignment with CLIP's shared anisotropic direction.
We next test whether alignment with this shared direction is causally related to concept loss.
To isolate the effect of CLIP's shared direction, we perturb each concept embedding by an exact norm-preserving rotation in the plane spanned by $c_i$ and $\hat{\mu}$, while fixing the displacement to $\|c_i'-c_i\|_2=\alpha\|c_i\|_2$.
Rotating the embedding toward $\hat{\mu}$ increases the concept loss rate from $27.3\%$ to $35.7\%$, whereas rotating it away reduces the rate to $22.3\%$. 
Rotations along random directions orthogonal to $c_i$, with the same displacement magnitude, produce negligible changes.
These results show that movement toward CLIP's shared direction increases concept-loss even when the perturbation magnitude is controlled.
\paragraph{Higher Entropy in Later Concepts.}
We next test whether token entropy exhibits the positional pattern observed in
rendering failures.
We construct semantically equivalent prompts with reversed concept order,
\texttt{"a $\langle$attr$_1\rangle$ $\langle$obj$_1\rangle$ and a
$\langle$attr$_2\rangle$ $\langle$obj$_2\rangle$"}
and
\texttt{"a $\langle$attr$_2\rangle$ $\langle$obj$_2\rangle$ and a
$\langle$attr$_1\rangle$ $\langle$obj$_1\rangle$"},
and additionally evaluate synthetically generated prompts containing three
attribute--object pairs.
As shown in Fig.~\ref{fig:fig2_analysis1}b, later-mentioned attributes and
objects consistently exhibit higher entropy.
Because some nouns in CC-500 occur preferentially at particular prompt
positions, the observed entropy difference could be confounded by vocabulary.
To control for this effect, we repeat the analysis using only nouns that
appear in both positions and compare their entropy when placed earlier versus
later in the prompt.
The later position entropy remains significantly higher for CLIP-G, with an
average increase of $0.0033$.
The standardized paired effect size is $d_z=1.08$, indicating a large
within concept positional effect, and the difference is statistically
significant ($p=1.9\times10^{-9}$).
These results show that the increase in entropy for later mentioned concepts
cannot be explained solely by differences in the vocabulary occupying each
prompt position.

\paragraph{Entropy as a Proxy for Concept Loss.}
Finally, we measure whether entropy is associated with actual rendering
failures.
Using the setting of Fig.~\ref{fig:fig2_analysis1}a, we compare token entropy
across correctly and incorrectly rendered concepts.
As shown in Fig.~\ref{fig:fig2_analysis1}c, for both attributes and objects,
entropy is higher when the requested concept is misaligned, mixed, or missing
than when it is correctly rendered.
The relationship also persists at the prompt level.
Mean concept entropy increases as rendering deteriorates from both concepts
being present to one or neither being present, and concept loss frequency rises
substantially across entropy quantiles.
At the individual-concept level, token entropy provides a graded signal of
susceptibility to concept loss rather than a discrete prediction.
We therefore use it to adapt the strength of embedding preservation for each
concept, assigning stronger intervention to representations with higher
entropy.
Additional predictive-validity, calibration, and causal-intervention results
are provided in the supplementary material.

\subsection{Token Relationships in L0 Attention}
\label{sec:analysis2}

Concept confusion requires identifying not only whether an individual token is weakly represented, but also how strongly an attribute is bound to its intended object.
We therefore analyze relational information within the CLIP text encoder.
Hereafter, a \emph{related pair} denotes the correct attribute--object
association, e.g.,
\texttt{$\langle$attr$_1\rangle$-$\langle$obj$_1\rangle$},
whereas an \emph{unrelated pair} associates the same attribute with an object
from another concept, e.g.,
\texttt{$\langle$attr$_1\rangle$-$\langle$obj$_2\rangle$}.
We find that L0 self-attention contains a useful signal of this relationship.
\begin{figure*}[t!]
  \centering
  \includegraphics[width=\textwidth]{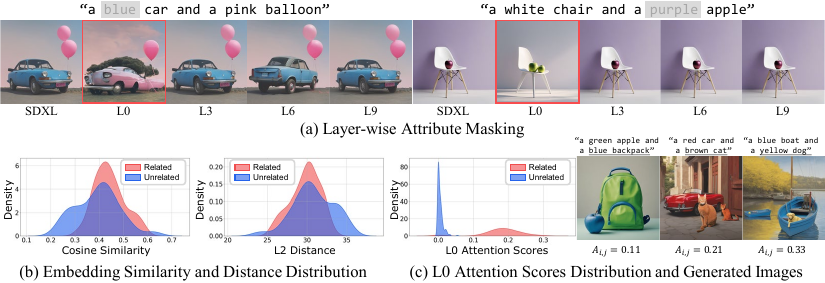}
  \caption{\textbf{Analysis of text embeddings and self-attention in the CLIP text encoder.}
(a) Layer-wise attribute masking shows that generation is particularly sensitive to attribute interventions at the first self-attention layer (L0), compared with later layers.
(b) Conventional embedding-space metrics show substantial overlap between related and unrelated attribute--object pairs.
(c) L0 attention provides a continuous signal of attribute--object relationships.}
  \label{fig:fig3_analysis2}
\end{figure*}
\vspace{-5pt}
\paragraph{Similarity Metrics Between Embeddings.}
We first ask whether conventional embedding space metrics reveal the correct
attribute--object association.
Using prompts from the A\&E dataset~\cite{ane}, we compare cosine similarity
and L2 distance for related and unrelated attribute--object pairs.
As shown in Fig.~\ref{fig:fig3_analysis2}b, the resulting distributions
substantially overlap.
Thus, the final token embeddings alone do not provide a clear measure of
binding strength.

\vspace{-5pt}
\paragraph{Layer-wise Self-Attention Analysis for Binding.}
Motivated by observations that different Transformer layers play distinct
functional roles~\cite{vig2019analyzing,vig2019multiscale,clark2019does}, we
instead examine self-attention throughout the CLIP text encoder.
We selectively mask the key of an attribute token at individual
self-attention layers and inspect its effect on generation.
As illustrated in Fig.~\ref{fig:fig3_analysis2}a, masking the attribute at L0
frequently removes or weakens the corresponding attribute in the generated
image, whereas applying the same intervention at later layers has much smaller
effect.
This indicates that early self-attention plays a particularly important role
in establishing compositional relationships.

We further test whether the effect of L0 attribute masking can be explained simply by the magnitude of the induced representation change.
Masking function words at L0 produces a conditioning-space displacement $8.4\times$ larger than masking the target attribute, yet increases the target concept-loss rate by only $1.5$ percentage points.
In contrast, masking the target attribute causes a much smaller representation change but increases the loss rate by $58.5$ percentage points.
The effect also depends strongly on layer depth, as applying the same attribute mask at a middle layer increases the loss rate by only $1.0$ percentage points.
Together, these results indicate that the large effect of L0 attribute masking arises from disrupting semantically relevant information at the first layer, supporting the role of L0 in establishing attribute--object binding.

\paragraph{L0 Attention as a Continuous Binding Signal.}
We next examine whether L0 attention between attribute and object tokens provides a continuous signal of their binding relationship.
As shown in Fig.~\ref{fig:fig3_analysis2}c, before controlling for token distance, related and unrelated pairs are strongly separable, with AUCs of $0.980$ for CLIP-L and $0.965$ for CLIP-G.
However, related attribute--object pairs are often closer in the prompt by construction, and attention may naturally be stronger between nearby tokens.
We therefore apply two controls for token distance.
First, we regress token distance out of the attention scores, yielding AUCs of $0.648$ for CLIP-L and $0.624$ for CLIP-G.
Second, we compare related pairs with unrelated pairs matched by token distance, obtaining AUCs of $0.634$ and $0.658$, respectively.
Although the separation decreases substantially after controlling for distance, L0 attention remains informative about the underlying attribute--object relationship.
Together with the controlled masking experiments above, these results indicate that L0 attention contains relationship-specific information beyond token proximity and is relevant to the resulting generation.
We therefore use the L0 attention score as a continuous, adaptive signal for determining the strength of attribute--object binding in IEB.

\begin{figure*}[t!]
  \centering
  \includegraphics[width=\textwidth]{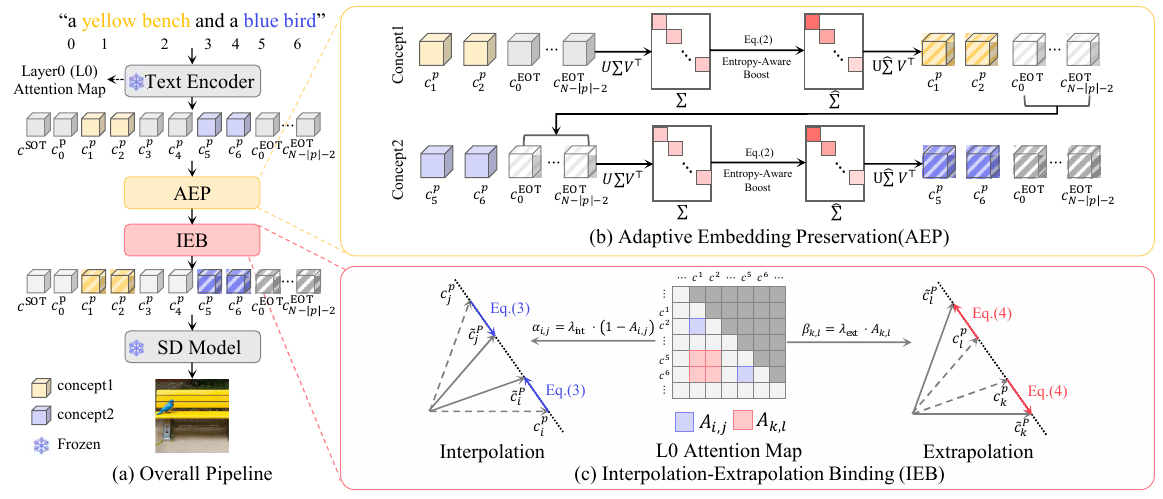}
  \caption{
  \textbf{Overview of the proposed TIE framework.}
  (a) Overall pipeline of TIE.
  (b) Adaptive Embedding Preservation (AEP), which decorrelates each
    concept from its preceding concepts and adaptively modifies text
    embeddings according to token entropy to mitigate concept loss.
  (c) Interpolation-Extrapolation Binding (IEB), which uses first-layer
  self-attention as a continuous binding signal to mitigate concept confusion.
  }
  \label{fig:fig3_pipeline}
\end{figure*}
\section{Methods}
\label{sec:methods}
Our method is motivated by the two key indicators derived from Section~\ref{sec:analysis}: \textit{token entropy} for concept loss and \textit{L0 attention score} for concept confusion.
We improve T2I generation in multi-object and multi-attribute scenarios by integrating two complementary components as shown in Fig.~\ref{fig:fig3_pipeline}.
To address concept loss, we first implement Adaptive Embedding Preservation (AEP), which exploits the tendency of tokens with higher token entropy. 
After recovering lost concepts, we mitigate concept confusion through Interpolation Extrapolation Binding (IEB), which estimates concept binding strength based on L0 attention scores.

\subsection{Adaptive Embedding Preservation}
\label{sec:eigenboost}
To preserve concepts that are more prone to concept loss using individual token embeddings, we introduce AEP. 
This method selectively amplifies concept token embeddings with end-of-text (EOT) tokens using entropy-aware singular value decomposition (SVD).
Prior work has shown that EOT tokens also encode semantic information crucial for concept composition~\cite{magnet, paddingtone}.
Our AEP jointly strengthens explicit concept token embeddings and EOT tokens according to their entropy, successfully mitigating concept loss.

\paragraph{Singular Value Decomposition for Concepts.}
SVD decomposes a matrix into orthogonal directions ordered by singular values, concentrating semantic information into its leading components.
Motivated by previous works~\cite{getwhat,oneprompt}, we leverage SVD to amplify singular values in concept token embeddings along with EOT tokens, thereby concentrating information for tokens with higher entropy to mitigate concept loss.

For simplicity, we explain with prompts in which only one attribute is paired with each object. For each attribute–object pair, we build the token-related matrix
\(
   X := 
   \bigl[\,c^{\text{attribute}}_k,\;
   c^{\text{object}}_k,
          c^{\text{EOT}}_0,\dots,
          c^{\text{EOT}}_{\,N-|\mathcal{P}|-2}\bigr],
\)
where $c^{\text{concept}}_k$ is the embedding of the $k$-th pair (Fig.~\ref{fig:fig3_pipeline}).  
We first gather the token embeddings
$
  c=\bigl\{c^{\text{SOT}},\;
  c^{\mathcal{P}}_{0},\dots,c^{\mathcal{P}}_{|\mathcal{P}|-1},\;
  c^{\text{EOT}}_{0},\dots,c^{\text{EOT}}_{N-|\mathcal{P}|-2}\bigr\},
$
where $c^{\text{SOT}}$ is the start-of-text (SOT) token embeddings, $c^{\mathcal{P}}_{i}$ is the $i$-th prompt token, and $c^{\text{EOT}}_{i}$ denotes the $i$-th EOT token. 
Decomposing $X$ gives
\(
   X = U \Sigma V^{\!\top},\quad
   \Sigma=\operatorname{diag}(\sigma_0,\sigma_1,\dots,\sigma_{n_0-1}),\;
   n_0=\min(M,N-|\mathcal{P}|-1)
\) where $U$ and $V$ are orthogonal matrices.
Each singular value is directly rescaled as
\(
    \hat{\sigma}_i=\sigma_i\,\delta\,e^{\gamma\sigma_i},
\) with $\gamma$ and $\delta$ as positive constants, concentrating the information in the concept embedding.
\paragraph{Proposition 1.} \textit{Let $\sigma_1\!\ge\!\cdots\!\ge\!\sigma_n\!>\!0$ and
$f(\sigma)=\sigma g(\sigma)$ with $g$ positive and increasing. Writing
$p_i=\sigma_i/\sum_j\sigma_j$ and $\hat p_i=f(\sigma_i)/\sum_j f(\sigma_j)$,
we have $\hat p \succ p$, hence
$\mathcal{H}(\hat p)\le\mathcal{H}(p)$ and $\mathrm{erank}(\hat X)\le\mathrm{erank}(X)$.\footnote{$\hat
p_i/p_i\propto g(\sigma_i)$ is non-increasing, giving majorization; Shannon entropy
is Schur-concave. Both $g(\sigma)=\delta e^{\gamma\sigma}$ here and
$g(\sigma)=\delta e^{\gamma\sigma\tilde{\mathcal{H}}}$ in Eq.~\eqref{eq:entropy_boost}
are increasing for $\gamma>0$.}}

Proposition~1 shows the rescaling provably concentrates the spectrum along a
small set of dominant directions. Empirically the effective rank drops by
$8$--$13\%$ ($4.28\!\to\!3.92$ on CLIP-L, $12.73\!\to\!11.67$ on CLIP-G), and
the token entropy of Eq.~\eqref{eq:token_entropy} moves in the same direction
for $99.9\%$ of concepts, so the two notions of concentration agree here.
However, preventing concept loss requires a more adaptive mechanism that can
predict the risk of omission for individual concepts.
\begin{table*}[t]
\setlength{\tabcolsep}{11pt}
\renewcommand{\arraystretch}{1}
\centering
\caption{
\textbf{Quantitative results of concept binding performance.}
All methods are evaluated with their corresponding backbones.
BLIP-VQA~\cite{huang2023t2i} scores are measured on five categories: Color, Shape, Texture, Spatial, and Complex prompts.
Inference time is compared across methods to assess efficiency.
Best results are shown in \textbf{bold} and second-best results are \underline{underlined}.
For SD3.5-L, only \textbf{bold} formatting is used due to the limited number of compared methods.
}
\label{tab:comparison}
\resizebox{\textwidth}{!}{
\begin{tabular}{llccccccc}
\toprule
\multirow{2}{*}[-0.4em]{Backbone}
  & \multirow{2}{*}[-0.4em]{Method}
  & \multirow{2}{*}[-0.4em]{\shortstack{Opt-\\free}}
  & \multirow{2}{*}[-0.4em]{\makecell{Inference\\Time (s) $\downarrow$}}
  & \multicolumn{5}{c}{BLIP-VQA $\uparrow$} \\
\cmidrule(lr){5-9}
  & & & & Color & Shape & Texture & Spatial & Complex \\
\midrule

SDXL
&                        &         & 9.82
& 0.5940
& 0.5068
& 0.5961
& 0.6249
& 0.4437 \\

& $\text{A\&E}_{XL}$     &         & 62.72
& 0.5934
& \underline{0.5082}
& \underline{0.6040}
& 0.6230
& 0.4434 \\

& $\text{SynGen}_{XL}$   &         & 22.42
& 0.5948
& 0.5059
& 0.5955
& \underline{0.6258}
& 0.4426 \\

& ToMe                   &         & 42.73
& 0.5161
& 0.3551
& 0.5621
& --
& 0.2818 \\

& Magnet                 & \cmarkg & \textbf{10.66}
& \underline{0.6833}
& 0.5076
& 0.5963
& 0.6239
& \underline{0.4507} \\

& \multicolumn{1}{>{\columncolor{lightgray}}l}{\textbf{TIE (Ours)}}
& \multicolumn{1}{>{\columncolor{lightgray}}c}{\cmarkg}
& \multicolumn{1}{>{\columncolor{lightgray}}c}{\underline{11.86}}
& \multicolumn{1}{>{\columncolor{lightgray}}c}{\textbf{0.7004}}
& \multicolumn{1}{>{\columncolor{lightgray}}c}{\textbf{0.5632}}
& \multicolumn{1}{>{\columncolor{lightgray}}c}{\textbf{0.6703}}
& \multicolumn{1}{>{\columncolor{lightgray}}c}{\textbf{0.6320}}
& \multicolumn{1}{>{\columncolor{lightgray}}c}{\textbf{0.4577}} \\

\midrule

SD3.5-M
&                        &        &10.33 
& 0.7838
& 0.5933
& 0.7457
& 0.6610
& 0.4972 \\

& FOCUS                  &         &45.55
& 0.7873
& 0.6071
& 0.7570
& \underline{0.6654}
& \underline{0.5025} \\

& TACA                   &         &{14.43}
& \underline{0.7912}
& \underline{0.6088}
& \underline{0.7584}
& \textbf{0.6671}
& \textbf{0.5126} \\

& Magnet                   & \cmarkg        &\underline{10.93}
& 0.7907
& 0.5976
& 0.7469
& 0.6610
& 0.4984 \\
& \multicolumn{1}{>{\columncolor{lightgray}}l}{\textbf{TIE (Ours)}}
& \multicolumn{1}{>{\columncolor{lightgray}}c}{\cmarkg}
& \multicolumn{1}{>{\columncolor{lightgray}}c}{\textbf{10.48}}
& \multicolumn{1}{>{\columncolor{lightgray}}c}{\textbf{0.7938}}
& \multicolumn{1}{>{\columncolor{lightgray}}c}{\textbf{0.6120}}
& \multicolumn{1}{>{\columncolor{lightgray}}c}{\textbf{0.7603}}
& \multicolumn{1}{>{\columncolor{lightgray}}c}{0.6586}
& \multicolumn{1}{>{\columncolor{lightgray}}c}{0.4986} \\

\midrule

SD3.5-L
&                        &         & 15.62
& 0.7766
& 0.6117
& 0.7518
& 0.6837
& 0.5050 \\

& \multicolumn{1}{>{\columncolor{lightgray}}l}{\textbf{TIE (Ours)}}
& \multicolumn{1}{>{\columncolor{lightgray}}c}{\cmarkg}
& \multicolumn{1}{>{\columncolor{lightgray}}c}{17.13}
& \multicolumn{1}{>{\columncolor{lightgray}}c}{\textbf{0.7826}}
& \multicolumn{1}{>{\columncolor{lightgray}}c}{\textbf{0.6185}}
& \multicolumn{1}{>{\columncolor{lightgray}}c}{\textbf{0.7548}}
& \multicolumn{1}{>{\columncolor{lightgray}}c}{\textbf{0.6933}}
& \multicolumn{1}{>{\columncolor{lightgray}}c}{\textbf{0.5154}} \\

\bottomrule
\end{tabular}
}
\end{table*}

\begin{table}[!t] 
\setlength{\tabcolsep}{3pt} 
\renewcommand{\arraystretch}{0.95} 
\centering 
\scriptsize 
\caption{ 
\textbf{Ablation study on T2I-CompBench.} 
AEP and IEB improve concept binding both individually and jointly on the SD3.5-Large backbone. 
} 
\resizebox{\columnwidth}{!}{ 
\begin{tabular}{c c c c c c c c} 
\toprule 
\multirow{2}{*}[-0.3em]{Exp. \#} &  
\multirow{2}{*}[-0.3em]{AEP} &  
\multirow{2}{*}[-0.3em]{IEB} &  
\multicolumn{5}{c}{BLIP-VQA $\uparrow$} \\ 
\cmidrule(lr){4-8} 
& & & Color & Shape & Texture & Spatial & Complex \\ 
\midrule 
 
1 &  &   
& 0.7766 & 0.6117 & 0.7518 & 0.6837 & 0.5050 \\ 
 
2 & \cmarkg &   
& 0.7775 & \underline{0.6167} & 0.7529 & 0.6931 & 0.5161 \\ 
 
3 &  & \cmarkg 
& \textbf{0.7831} & 0.6137 & \textbf{0.7550} & \underline{0.6923} & \textbf{0.5171} \\ 
 
4 (Ours) & \cmarkg & \cmarkg 
& \underline{0.7826} & \textbf{0.6185} & \underline{0.7548} & \textbf{0.6933} & \underline{0.5154} \\ 
 
\bottomrule 
\end{tabular} 
} 
\label{tab:tab3_ablation_ieonly} 
\end{table}

\paragraph{Entropy-aware Boost.}
As shown in Sec.~\ref{sec:analysis1}, token entropy provides a continuous proxy for concept-loss propensity.
Accordingly, we apply entropy-aware scaling that assigns stronger amplification to higher-entropy concepts rather than uniformly rescaling all embeddings.
We compute the token entropy $\mathcal{H}_i$ as in Eq.~\eqref{eq:token_entropy} and average it over the $N-|\mathcal{P}|$ embeddings in the concept-specific matrix $X$:
\begin{equation*}
    \tilde{\mathcal{H}}
    =
    \frac{1}{N-|\mathcal{P}|}
    \sum_{i=1}^{N-|\mathcal{P}|}
    \mathcal{H}_i .
\end{equation*}
Let $\sigma_i$ denote the $i$-th singular value of $X$ obtained from
$X=U\Sigma V^\top$.
Since later concepts inherit information from earlier ones under
causal attention, we first decorrelate each subsequent concept by
subtracting $C$ times the mean embedding of its preceding concepts,
$\bar{c}_k^{\mathcal{P}} = c_k^{\mathcal{P}} - \frac{C}{k-1}\sum_{j<k}
c_j^{\mathcal{P}}$, and construct $X$ from the resulting embeddings.
We rescale each original singular value as
\begin{equation}
\label{eq:entropy_boost}
    \hat{\sigma}_i
    =
    \sigma_i \cdot \delta
    \exp\!\left(
        \gamma \sigma_i \tilde{\mathcal{H}}
    \right),
\end{equation}
where $\gamma>0$ controls the strength of entropy-aware amplification and
$\delta>0$ is a global scaling factor.
Consequently, concept representations with higher entropy receive stronger
spectral concentration, while lower-entropy concepts are modified more conservatively.

\subsection{Interpolation-Extrapolation Binding}
\label{sec:iebinding}

\begin{figure*}[t!]
  \centering
    \includegraphics[width=\textwidth]{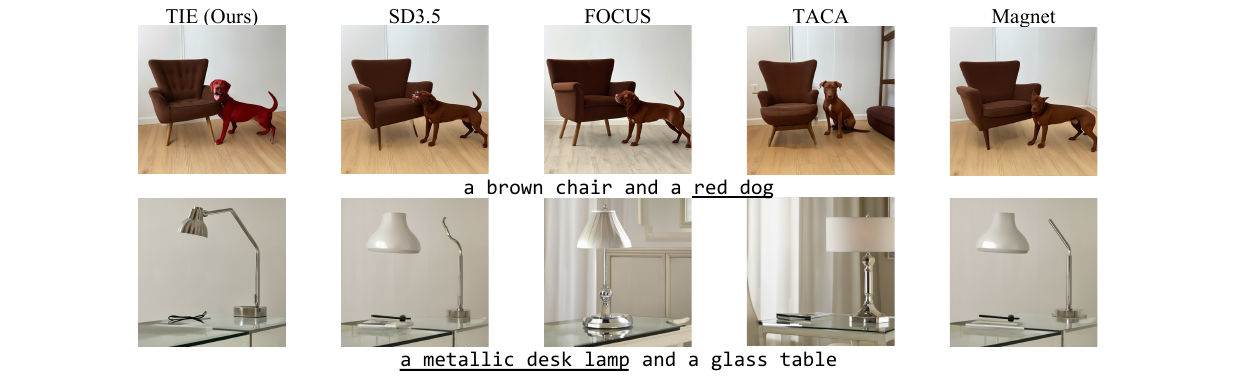} 
\caption{\textbf{Qualitative comparison of TIE with SD3.5-based models.} We compare TIE with other SD3.5-based models. The results show that TIE better preserves all objects, along with their spatial relationships and colors, in alignment with the given prompt.
}
\label{fig:fig5_qualitative}
\end{figure*}
After addressing concept loss through AEP, we now propose the Interpolation–Extrapolation Binding (IEB) mechanism to resolve concept confusion, ensuring that the relationships between concepts are accurately reflected in the generated image, as illustrated in Fig.~\ref{fig:fig3_pipeline}c.
Inspired by prior studies indicating that CLIP text embeddings exhibit compositional structure~\cite{embeddingspace}, we incorporate concept tokens to both attribute and object tokens bidirectionally to bring their embeddings closer.
In contrast, we extrapolate unrelated concepts by pushing them further apart. 
To reflect binding strength between tokens, IEB leverages the \textit{L0 attention score} introduced in Section~\ref{sec:analysis2}.
\paragraph{Interpolation for Binding.}
We strengthen the association between related attribute--object pairs
by interpolating their embeddings toward each other.
Let $(c^{\mathcal{P}}_i,c^{\mathcal{P}}_j)$ denote a related token pair,
where $c^{\mathcal{P}}_i$ corresponds to the attribute and
$c^{\mathcal{P}}_j$ to its object.
Their updated embeddings are
\begin{equation}
\footnotesize
\tilde{c}^{\mathcal{P}}_i
=
(1-\alpha_{i,j})c^{\mathcal{P}}_i
+
\alpha_{i,j}c^{\mathcal{P}}_j,
\qquad
\tilde{c}^{\mathcal{P}}_j
=
(1-\alpha_{i,j})c^{\mathcal{P}}_j
+
\alpha_{i,j}c^{\mathcal{P}}_i.
\label{eq:interpolation}
\end{equation}
Here, $\alpha_{i,j}=\lambda_{\mathrm{int}}(1-A_{i,j})$, where $A_{i,j}$
is the L0 attention score between the pair.
Thus, weakly bound pairs receive stronger interpolation, while
$\lambda_{\mathrm{int}}$ controls the overall update strength.

\paragraph{Extrapolation for Repulsion.}
After reinforcing related pairs, we suppress interference between unrelated
concepts by extrapolating their embeddings apart.
For an unrelated pair $(c^{\mathcal{P}}_k,c^{\mathcal{P}}_l)$, we update
\begin{equation}
\footnotesize
\tilde{c}^{\mathcal{P}}_k
=
(1+\beta_{k,l})c^{\mathcal{P}}_k
-
\beta_{k,l}c^{\mathcal{P}}_l,
\qquad
\tilde{c}^{\mathcal{P}}_l
=
(1+\beta_{k,l})c^{\mathcal{P}}_l
-
\beta_{k,l}c^{\mathcal{P}}_k.
\label{eq:extrapolation}
\end{equation}
Here, $\beta_{k,l}=\lambda_{\mathrm{ext}}A_{k,l}$.
Thus, unrelated concepts with stronger attention receive stronger repulsion,
while $\lambda_{\mathrm{ext}}$ controls the overall update strength.

\section{Experiment}
\label{sec:experiment}
\paragraph{Baselines.}
We compare our model, TIE, with baselines including A\&E~\cite{ane}, SynGen~\cite{syngen}, Magnet~\cite{magnet}, and ToMe~\cite{tome}, all of which are built on the SDXL~\cite{sdxl} backbone.
For fair comparison, we adapt A\&E and SynGen to the SDXL base model, since their original implementations are not designed for it.
When using SD3.5-Medium as the backbone, we compare against FOCUS~\cite{bill2026focus} and TACA~\cite{lv2025rethinking}.
For SD3.5-Large, we report the improvement over the corresponding backbone
to assess the scalability of TIE to a larger model.

\vspace{-5pt}
\paragraph{Evaluation Metrics.}
We employ the BLIP-VQA metric from T2I-CompBench~\cite{huang2023t2i} for evaluation.
This metric utilizes BLIP for visual question answering, enabling fine-grained
assessment of attributes in the generated images.
Following prior works~\cite{tome,teemo,ranni,ella,comat}, we evaluate the
Color, Shape, and Texture categories, and further include the Complex and
Spatial categories for a more comprehensive evaluation.
For each category, we evaluate all 300 benchmark prompts and generate five
images per prompt using five fixed random seeds.
Inference time is measured on an RTX A6000 GPU with a batch size of 1.
Additional results using other metrics and datasets are provided in the
supplementary material.

\paragraph{Implementation Details.}
We use Stanza to parse each prompt and identify its attribute--object
relations. L0 attention scores are averaged across attention heads, and for
concepts spanning multiple CLIP tokens, the corresponding token-level
quantities are averaged over their constituent tokens.

\subsection{Results}
\paragraph{Quantitative Comparison.}
As shown in Tab.~\ref{tab:comparison}, we compare TIE with baselines built on the SDXL and SD3.5 backbones.
For SDXL, our method achieves superior performance across all categories compared to other baseline models.
We exclude ToMe~\cite{tome} from the spatial category evaluation as it is specifically designed for attribute binding and does not incorporate spatial structure in its token merging process.
For SD3.5-Medium, TIE achieves the best performance on Color, Shape, and Texture, outperforming recent baselines including FOCUS and TACA, while remaining competitive on Spatial and Complex prompts.
TIE also achieves strong performance on SD3.5-Large, demonstrating consistent gains across model scales.
Overall, these results demonstrate that our method effectively mitigates concept confusion and prevents concept loss with minimal overhead.

\paragraph{Qualitative Comparison.}
We qualitatively compare TIE with existing semantic binding methods, as shown in Fig.~\ref{fig:fig5_qualitative}.
The results demonstrate that TIE achieves improved semantic binding performance.
In Fig.\ref{fig:fig5_qualitative}, applying our method to SD3.5 alleviates concept loss while also improving spatial consistency compared to the baseline models.
These results highlight the enhanced robustness of TIE in preserving both attribute and object integrity during generation.
Additional qualitative examples on diverse models are provided in the supplementary material.

\section{Conclusion}
\label{sec:conclusion}
Text-to-image diffusion models struggle with multi-concept prompts.
We find that token entropy predicts concept loss, while L0 attention captures attribute–object binding.
We propose TIE, a training-free framework combining entropy-aware singular value amplification (AEP) and L0-attention-based binding (IEB) in a single embedding update.
TIE mitigates concept loss and confusion, improving compositional alignment across SDXL and SD3.5 without training or model modifications.

\section*{Acknowledgments}
This work was supported in part by the IITP RS-2024-00457882 (AI Research Hub Project), IITP 2020-II201361, NRF RS-2024-00345806, NRF RS-2023-002620, and IITP RS-2026-25616313.
Affiliations: Department of Artificial Intelligence (S.H, T.E.C, Y.J, C.Y, S.J.H).
{
    \small
    \bibliographystyle{ieeenat_fullname}
    \bibliography{main}
}

\end{document}